\documentclass[twocolumn]{autart}    

\usepackage{graphicx}          
\usepackage{hyperref}
\usepackage{amsmath}
\usepackage{mathtools}
\usepackage{xcolor}
\usepackage{bbm}
\usepackage{algorithm}
\usepackage{algpseudocode}
\usepackage{cite}
\usepackage{mathtools}
\usepackage{grffile}
\usepackage{multirow}
\usepackage{diagbox}
\usepackage{physics}
\renewcommand{\vec}{{\bf vec}}
\usepackage{amssymb}

\DeclareMathOperator*{\argmin}{arg\,min}
\usepackage[overload]{empheq}
\begin{document}

\begin{frontmatter}

\title{DeepBayes - an estimator for parameter estimation in stochastic nonlinear dynamical models} 

\thanks[footnoteinfo]{This work was supported by the Digital Futures center (KTH), VINNOVA Competence Centre AdBIOPRO, contract 2016-05181, and by the Swedish Research Council through the project 2019-04956, and the research environment NewLEADS - New Directions in Learning Dynamical Systems, contract 2016-06079.}

\author[Stockholm1]{Anubhab Ghosh}\ead{anubhabg@kth.se},    
\author[Uppsala]{Mohamed Abdalmoaty}\ead{mohamed.abdalmoaty@it.uu.se},
\author[Stockholm1]{Saikat Chatterjee}\ead{sach@kth.se}, 
\author[Stockholm2]{H\aa kan Hjalmarsson}\ead{hjalmars@kth.se},               

\address[Stockholm1]{Division of Information Science and Engineering and Digital Futures, KTH Royal Institute of Technology, Stockholm, Sweden}  
\address[Uppsala]{Division of Systems and Control, Department of Information Technology, Uppsala University, Uppsala, Sweden}             
\address[Stockholm2]{Division of Decision and Control Systems, Digital Futures, and the Competence Centre AdBIOPRO, KTH Royal Institute of Technology, Stockholm, Sweden}        


\begin{abstract}                          
Stochastic nonlinear dynamical systems are ubiquitous in modern, real-world applications. Yet, estimating the unknown parameters of stochastic, nonlinear dynamical models remains a challenging problem. The majority of existing methods employ maximum likelihood or Bayesian estimation. However, these methods suffer from some limitations, most notably the substantial computational time for inference coupled with limited flexibility in application. In this work, we propose DeepBayes estimators that leverage the power of deep recurrent neural networks in learning an estimator. The method consists of first training a recurrent neural network to minimize the mean-squared estimation error over a set of synthetically generated data using models drawn from the model set of interest. The \textit{a priori} trained estimator can then be used directly for inference by evaluating the network with the estimation data. The deep recurrent neural network architectures can be trained offline and ensure significant time savings during inference. We experiment with two popular recurrent neural networks -- long short term memory network (LSTM) and gated recurrent unit (GRU). We demonstrate the applicability of our proposed method on different example models and perform detailed comparisons with state-of-the-art approaches. We also provide a study on a real-world nonlinear benchmark problem. The experimental evaluations show that the proposed approach is asymptotically as good as the Bayes estimator.
\end{abstract}

\end{frontmatter}

\section{Introduction}\vspace{-0.5em}
System identification of dynamical systems is a topic of great interest. In particular, system identification of linear dynamical systems has been studied extensively during the last six decades \cite{ljung2010perspectives, pintelon2012system, sjoberg1995nonlinear}, and this has led to the development of several useful toolboxes suited to this task \cite{ninness2013unit, ljung1988system}.

The identification problem for nonlinear dynamical systems is still a challenge. The presence of nonlinear transformations on the hidden variables leads to significant difficulty in directly applying conventional methods such as maximum likelihood (ML) estimation. 

Previous work on ML-based estimation include \cite{ghahramani1999learning, caines1974maximum, schon2011system, lindsten2013efficient}. One notable solution to the problem of estimating parameters using ML-based estimation involves the use of Monte Carlo expectation maximization (MCEM) \cite{wei1990monte} in combination with a particle smoother (PS) algorithm \cite{schon2011system}. This method is offline, meaning that first data are collected and then the unknown parameters are estimated. Furthermore, it can cater to a general class of discrete-time, nonlinear dynamical systems. For succinctness, we abbreviate this method as PSEM. Although offline sampling-based methods like PSEM, are quite versatile, they are computationally expensive. The success of the solution in \cite{schon2011system} requires that the number of particles used should increase with the number of iterations of the optimization algorithm in order to attain convergence. In order to tackle this issue, another offline ML-based estimation algorithm was proposed in \cite{lindsten2013efficient}. This algorithm uses a combination of Markovian, stochastic approximation expectation maximization (referred to as SAEM) and a conditional particle filter (CPF) with ancestral sampling. The CPF-SAEM method reduces the estimation time but it may not be more accurate compared to the PSEM method. 

Another method involves the use of linear prediction error methods (PEM) for stochastic nonlinear models \cite{abdalmoaty2019linear,abdalmoaty2020identification}. This approach relies on the statistics of the model, involves rather inexpensive computations (since they do not require exact computation of the likelihood) and are yet highly competitive with respect to sequential Monte Carlo (SMC) based methods similar to \cite{lindsten2013efficient}. A downside is that such estimators are statistically, asymptotically inefficient. 

An alternative approach to solving the estimation problem involves using a Bayesian framework, where practitioners make use of prior knowledge to improve the estimates  \cite{ninness2010bayesian}. This approach can help in obtaining an optimal estimate, e.g. using the conditional mean estimator (CME). However, the computations involve calculation of normalization constants that require solving analytically intractable integrals. A way to avoid this predicament is to use numerical approximation methods such as Markov Chain Monte Carlo \cite{ninness2010bayesian, kantas2015particle, andrieu2004particle}. 

The above methods are undoubtedly quite interesting, however it is evident that they have certain limitations. To elucidate on this, recall that an estimator is a map (function) from $\mathbb{R}^N$, where $N$ is the dimension of the data vector, to $\mathbb{R}^d$,  where $d$ is the dimension of the parameter vector. The prevailing paradigm in parameter estimation is to construct this map only for the point  in $\mathbb{R}^N$ which corresponds to the data vector for which the parameter estimate is desired. This is done at the time of the estimation, making the computation time for the training (the time it takes to construct the estimator) and the computation time for inference (the time it takes to compute the map from the given data vector to the parameter estimate) the same. Methods such as PSEM, CPF-SAEM and CME require the real data during the learning process; training time and inference time for such methods is the same. 
Moreover, such methods are susceptible to poor initialization and can get stuck in a poor local optimum.
CME also requires careful tuning of the proposal densities. This leads to the search for an alternative framework that can provide competitive estimates similar to Bayesian methods like CME but also enable faster inference. 

In this paper, we make a fundamental change of perspective by separating the construction of the estimator and the inference step. By constructing the entire estimator beforehand, the inference step can be done simply by evaluating the estimator at the point in $\mathbb{R}^N$ corresponding to the data vector of interest, making this step very simple and computationally fast. The construction of the estimator is done in a sampling-based, Bayesian fashion. It involves using an extensive number of synthetic data sets generated by simulating the models drawn from the model set of interest. For the estimator itself, we use a recurrent neural network (RNN) \cite{rumelhart1986learning} leading to what one could call a deep Bayesian estimator. We will refer to this class of estimators as DeepBayes estimators.

The use of neural networks in system identification is not new. Feedforward neural networks were used to construct predictors, i.e. inferring the relationship between past input-output data and predicting future outputs given a new input \cite{sjoberg1994neural}. For the output prediction, there exist a recent work where deep feedforward neural networks is used \cite{zancato2021novel}. Instead of output prediction, we address a different problem - estimating parameters of a nonlinear dynamical system using deep RNNs.

\subsection{Contributions}\vspace{-0.7em}
In this work, we tackle the problem of learning a DeepBayes estimator for the task of parameter estimation in stochastic, nonlinear dynamical models by leveraging the modeling capability of deep RNNs. The proposed methodology is general, and applies for any dynamical model that can be simulated. Hereto, we focus on discrete-time state-space and construct parameter estimators for discrete-time dynamical models that can be simulated. Our main contributions are as follows: We propose the DeepBayes estimation procedure where we employ deep RNNs for learning estimators. Using a mean squared error loss function, we demonstrate the performance of our proposed method and compare with state-of-the-art ML-based methods \cite{lindsten2013efficient, schon2011system} and the conditional mean estimator (CME) that uses  Metropolis-Hastings (MH) algorithm \cite{hastings1970monte, chib1995understanding, robert1999metropolis}. We illustrate the advantages of our proposed approach through simulation experiments using nonlinear state space models and on a real-world, nonlinear benchmark problem. 
\subsection{Paper outline}\vspace{-0.7em}
The paper is organized as follows: In Section \ref{sec:prob_statement}, we briefly describe the problem that we wish to solve. In Section \ref{sec:deepRNN}, a quick overview on deep RNNs is provided to help setting the stage for describing the architectures used. In Section \ref{sec:proposed_approach}, we introduce the DeepBayes estimators and provide theoretical motivations. Section \ref{sec:simulations} describes our experimental setup, training strategy, hyperparameter tuning and obtained results using the DeepBayes approach. We also describe the implementation of the CME using MH and compare with the CPF-SAEM and PSEM methods. Finally, in Section \ref{sec:conclusion}, we provide final comments related to our method and describe possible directions of future work. 
\subsection{Notations}\vspace{-0.5em}
In this paper, we use bold font lowercase symbols to denote vectors, and regular lowercase font to denote scalars, e.g. $\boldsymbol{x}$ represents a vector while $x_{i}$ represents the $i^{th}$ component of $\boldsymbol{x}$. Upper case symbols in bold font represent matrices. The symbol $\mathcal{M}$ is used to denote a model under consideration. The notation $\lbrace \cdot \rbrace$ is used to compactly define a sequence of signal values, e.g. $\lbrace y_k \rbrace_{k=1}^{N} $ denotes a sequence of signal values $\left[y_1, y_2, \ldots, y\!_N\right]$ with the $k^{th}$ signal value given by $y_k$. The symbol $\mathbf{E}\left[\cdot\right]$ represents the expectation operator, $\mathcal{L}\left(\cdot\right)$ denotes the Laplace transform, $\odot$ denotes element-wise multiplication. Uniform and Gaussian distributions have been represented by $\mathcal{U}\left[\cdot\right]$ and $\mathcal{N}\left(\cdot\right)$ respectively.

\section{Problem formulation}\label{sec:prob_statement}
We assume that we are provided with a data set of inputs and outputs of a dynamical system represented by
\begin{equation*}
    D\!_{N} := \lbrace \left(y_{k}, u_{k}\right): k= 1, 2, \ldots, N \rbrace, 
\end{equation*}
consisting of $N$ values of the output signal $\lbrace y_k \rbrace_{k=1}^{N}$ and input signal $\lbrace u_k \rbrace_{k=1}^{N}$. We assume that the model of interest $\mathcal{M}$ has the following general structure 
\begingroup\makeatletter\def\f@size{9.5}\check@mathfonts
\begin{equation}
\mathcal{M}:
\begin{cases}
\begin{aligned}
x_{k+1} &= f\left(x_{k}, u_{k}, w_{k}; \boldsymbol{\theta}\right), \\
y_{k} &= g\left(x_{k}, u_k, v_{k}; \boldsymbol{\theta} \right), \hspace{0.1 in} \left(k=1, \ldots, N\right),  \\
\boldsymbol{\theta} &\sim \pi\left(\boldsymbol{\theta}\right)
\end{aligned}
\end{cases}
\label{eq: model_structure}
\end{equation}
\endgroup
where $\boldsymbol{\theta} \in \Theta \subseteq  \mathbb{R}^{d}$ is an unknown parameter vector with a prior distribution $\pi\left(\boldsymbol{\theta}\right)$, $ x_{k}$ denotes a latent variable, $x_{1} = 0$,  and $w_{k}, v_{k}$ are unobserved noise processes whose distributions have  known forms and may be parameterized by $\boldsymbol{\theta}$, e.g. noise variances may be included in $\boldsymbol{\theta}$. 
The problem is to learn an estimator
$\hat{\varTheta}_{\boldsymbol{\alpha}}^{\mathcal{M}} \left(\cdot; \beta \right) : \mathbb{R}^N \to \mathbb{R}^d$ and obtain an estimate of the unknown parameter vector $\boldsymbol{\theta}$ as
\begin{equation}
    \hat{\boldsymbol{\theta}}(D\!_N) :=\hat{\varTheta}_{\boldsymbol{\alpha}}^{\mathcal{M}}    \left(\mathbf{Y}\!_N; \beta\right),
\label{eq:estimator_definition}
\end{equation}
where $\mathbf{Y}\!_N = \{y_k\}_{k=1}^{N}$ is the output signal of length $N$ given in $D\!_N$. The superscript $\mathcal{M}$ indicates the dependence on the model set and 
the subscript $\boldsymbol{\alpha}$ indicates the set of free parameters in the estimator that are learned when the estimator is constructed, e.g. weights and biases of the deep RNN in use. The parameter $\beta$ indicates the tunable hyperparameters such as learning rate $\eta$, number of epochs $N_{epochs}$, etc.  
However, it is worthwhile to note that our estimation approach could also be used with functional approximators other than RNNs. Another important fact is that in \eqref{eq:estimator_definition}, we have not included $\mathbf{U}\!_{N} = \{u_k\}_{k=1}^{N}$ since we consider the case where the input signal is fixed a priori, as often is the case in process industry \cite{ljung2010perspectives, zhu2001multivariable}, or not present at all, as in time series forecasting \cite{hewamalage2021recurrent}. 
\begin{figure}[!htbp]
    \centering
    \includegraphics[width=0.42\textwidth]{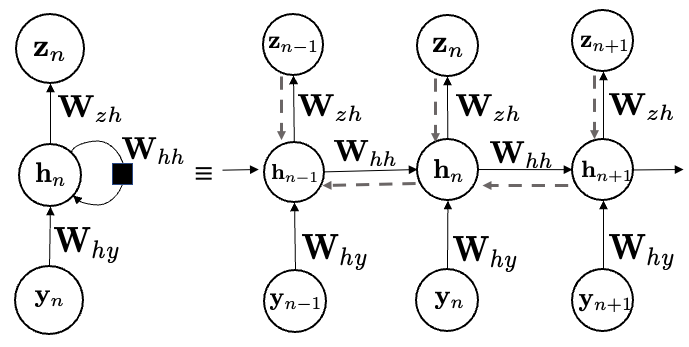}
    \caption{Illustration of a single RNN unit \textit{(left)} and illustration of the architecture as it \textit{unfolds} in time \textit{(right)}. The dashed line indicates the direction of gradient flow (backpropagation through time) in training \cite[Chap. 6]{goodfellow2016deep}}
    \label{fig:rnn_unfolding}
\end{figure}
\section{Deep recurrent neural networks}\label{sec:deepRNN}
We will employ RNNs having the structure shown in Fig. \ref{fig:rnn_unfolding}. Here, the input to the RNN is $\lbrace \mathbf{y}_n\rbrace_{n=1}^{N}$ where for generality we assume that $\mathbf{y}_n$ can be vector valued, and $\lbrace \mathbf{z}_n\rbrace_{n=1}^{N}$ is the output of the RNN.

The forward propagation in case of a single RNN unit can be represented by the following equation
\begin{equation}\label{eq:RNNeqn}
\begin{aligned}
\mathbf{h}_n &= \tanh \left(\mathbf{W}_{hh} \mathbf{h}_{n-1} + \mathbf{W}_{hy} \mathbf{y}_{n} + \mathbf{b}_h \right), \\
\mathbf{z}_n &= \phi\left(\mathbf{W}_{zh} \mathbf{h}_n + \mathbf{b}_{z} \right), 
\end{aligned}
\end{equation}
where $\mathbf{z}_n$ is the output at the $n^{th}$ time step, $\mathbf{h}_n$ is the hidden state of the RNN at the $n^{th}$ time step, $\tanh(x) = \frac{e^x - e^{-x}}{e^{x} + e^{-x}}$ denotes the hyperbolic tangent function and $\phi(\cdot)$ denotes a nonlinear activation function such as softmax or rectified linear unit (ReLU) \cite[Chap. 6]{goodfellow2016deep}. The weight matrices $\mathbf{W}_{hh}, \mathbf{W}_{hy}, \mathbf{W}_{zh}$ and bias vectors $\mathbf{b}_{h}, \mathbf{b}_{z}$ are learnable parameters of the RNN and are the same across time steps when the RNN is unfolded as shown in Fig. \ref{fig:rnn_unfolding}.

For classification tasks, a loss function such as cross-entropy is used for training, while for regression tasks, one often uses the mean squared error. The parameters in \eqref{eq:RNNeqn} are optimized using backpropagation through time (BPTT) \cite{rumelhart1986learning}. RNNs can even be stacked on top on one another to constitute a deep RNN \cite{pascanu2014construct}. 

RNNs are  quite versatile and possess several processing modes - one-to-one, one-to-many, many-to-one or many-to-many \cite[Chap. 10]{goodfellow2016deep}.  
In a many-to-one mode, RNNs are used for summarization of sequential data and used in applications such as word sentiment analysis \cite{tang2015document}. 
The RNN processes the entire length of the sequence and then uses the hidden state at the final time step $\mathbf{h}_N$ to project it nonlinearly to the desired output space as\vspace{-0.3em}
\begin{equation*}
\begin{aligned}
\mathbf{z}_N &= \phi\left(\mathbf{W}_{zh}\mathbf{h}_N + \mathbf{b}_z\right), \\
\hat{\boldsymbol{\theta}} &= \mathbf{W}_{\theta z}\mathbf{z}_N + \mathbf{b}_{\theta}, \\
\end{aligned}
\label{eq:dense_layers}
\end{equation*}
where, the matrices $\mathbf{W}_{zh} \in \mathbb{R}^{n_z \times n_H}, \mathbf{W}_{\theta z} \in \mathbb{R}^{d \times n_z}$ and biases $\mathbf{b}_{z} \in \mathbb{R}^{n_z \times 1}, \mathbf{b}_{\theta} \in \mathbb{R}^{d \times 1}$ are learned by the algorithm, $\phi(\cdot)$ is an element-wise nonlinearity such as ReLU function \cite{goodfellow2016deep}.

\textbf{LSTM and GRU:} One notable issue for BPTT is the vanishing gradient problem in the computation of the gradients. RNN training algorithms may become inefficient owing to either exploding or vanishing gradients. Long short term memory units (LSTMs) try to solve this problem by imposing gates which control the information flow \cite{hochreiterLongShortTermMemory1997}. Gated recurrent units (GRUs) provide some simplification over the conventional LSTM units and have a smaller number of gates \cite{choPropertiesNeuralMachine2014}. We refer the interested reader to \cite{chen2016gentle,karpathy2015visualizing} for reviews and visualization insights concerning these architectures. 

In the proposed DeepBayes method, we use the LSTM and GRU architectures in many-to-one mode and described in Section \ref{sec:simulations}.

\section{Proposed approach}
\label{sec:proposed_approach}
We assume that the user may a priori design an input signal to excite the system for identification purposes. Then, the objective is to learn an estimator for a given model structure and the chosen input signal. Given a model $\mathcal{M}$ as in \eqref{eq: model_structure}, we require a supervised learning setup to train the RNN in the proposed DeepBayes. This requires a training dataset. 

\begin{figure*}
\centering
  \includegraphics[width=0.7\textwidth]{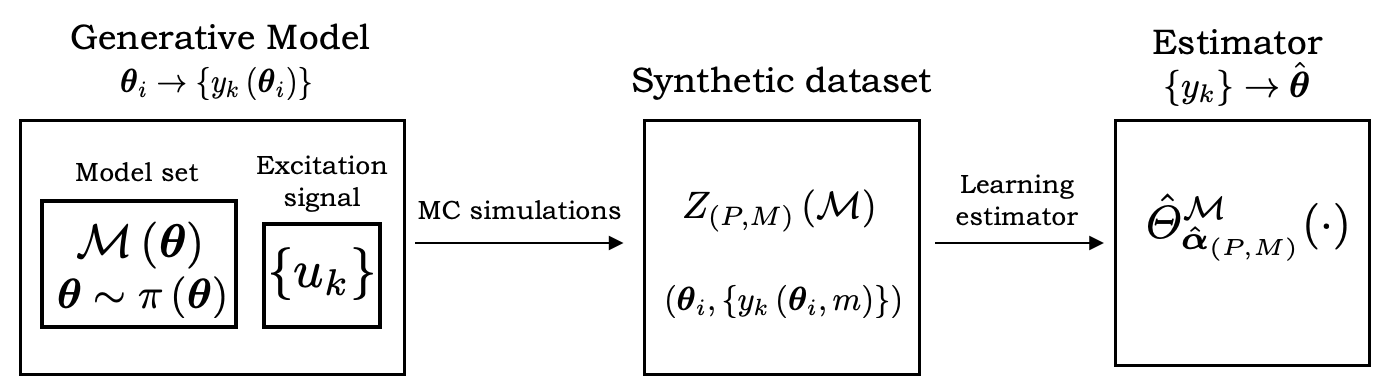}
  \caption{A schematic illustration of the proposed DeepBayes estimation procedure, where we generate the synthetic data set $Z_{(P, M)}\left(\mathcal{M}\right)$ through Monte Carlo simulation (denoted by `MC simulations') and learn the estimator model $\hat{\varTheta}^{\mathcal{M}}_{\hat{\boldsymbol{\alpha}}_{(P,M)}} (\cdot)$.} 
  \label{fig:procedure}
\end{figure*}

Our idea to create such a training dataset is to first generate a large number of output signals using parameters selected from a set $\Theta$ in the parameter space. We achieve this by first sampling $P$ parameter vectors $\lbrace \boldsymbol{\theta}_p \rbrace_{p=1}^{P}$ from the prior distribution $\pi\left(\boldsymbol{\theta}\right)$. For each sampled $\boldsymbol{\theta}_p$, we then generate $M$ signals consisting of noise realizations $w_{k}\left(m\right), v_{k}\left(m\right)$ for $m=1, 2, \ldots, M$. Lastly, we use $w_{k}\left(m\right), v_{k}\left(m\right)$ to generate  output signals $y_{k}\left(m\right)$ for $m=1, 2, \ldots, M$ according to \eqref{eq: model_structure}. In total, this gives $P \times M$ output signals denoted by
\begin{equation*}
\begin{aligned}
\mathbf{Y}\!_N(\boldsymbol{\theta}_p,m) &= \lbrace y_{k}\left(\boldsymbol{\boldsymbol{\theta}}_p, m\right) \rbrace_{k=1}^{N},
\end{aligned}
\end{equation*}
 and constitutes our synthetic training set 
\begin{equation*}
\begin{aligned}
Z_{(P, M)}\left(\mathcal{M}\right) &= \lbrace \left(\mathbf{Y}\!_N(\boldsymbol{\theta}_p,m), \boldsymbol{\theta}_p\right) \rbrace,
\end{aligned}
\label{eq:training_data}
\end{equation*}
where $p=1, 2, \ldots, P,\,  m=1, 2, \ldots, M$. 
Our goal is to learn an estimator $\hat{\varTheta}^{\mathcal{M}}_{\boldsymbol{\alpha}_{(P,M)}}(\cdot; \beta)$ by fitting an RNN to the synthetic data set $Z_{(P,M)}(\mathcal{M})$.
This involves tuning hyperparameters $\beta$ and learning the unknown parameters $\boldsymbol{\alpha}$ in the RNN. 
We do so by minimizing the mean squared error (MSE) loss function between the parameter estimates given by the estimator $\hat{\varTheta}^{\mathcal{M}}_{\boldsymbol{\alpha}}(\cdot; \beta)$ and the sampled parameter vectors. Thus, we compute
 \begingroup\makeatletter\def\f@size{9}\check@mathfonts
\begin{equation}
    \hat{\boldsymbol{\alpha}}_{(P,M)} = \arg\min_{\boldsymbol{\alpha}} \sum_{p=1}^P \sum_{m=1}^M\|\boldsymbol{\theta}_p  - \hat{\varTheta}_{\boldsymbol{\alpha}}^{\mathcal{M}} (\mathbf{Y}\!_N(\boldsymbol{\theta}_p,m); \beta)\|^2,
\label{eq:alpha_parameters}
\end{equation}
\endgroup
where $\hat{\boldsymbol{\alpha}}_{(P,M)}$ depends on $P$, the number of parameter realizations that is used, and $M$, the the number of output signals generated per parameter realization. A schematic illustration of this method is given in Fig. \ref{fig:procedure}.
\subsection{Theoretical motivation}
\label{sec:theoretical_results}
The learning problem is defined in \eqref{eq:alpha_parameters}. If the function $\hat{\varTheta}_{\hat{\boldsymbol{\alpha}}_{(P,M)}}^{\mathcal{M}}$ is flexible enough, i.e. the number of hidden layers, the number of units in each layer is sufficiently large, and the number of used samples $P$ and $M$ are sufficiently large,  the obtained estimator will approximately be the Bayes estimator corresponding to the MSE loss function. To see this, let us denote the conditional distribution of the model outputs 
by $p(\mathbf{Y}\!_N|\boldsymbol{\theta})$.  
Then, under mild conditions on the moments of $\boldsymbol{\theta}$ and $\mathbf{Y}\!_N$, we have
 \begingroup\makeatletter\def\f@size{9.5}\check@mathfonts
\begin{equation}\label{eq:BayesEstimator}
\hat{\boldsymbol{\alpha}}_{(P,M)} \to \arg\min_{\boldsymbol{\alpha}} \mathbf{E}_{p(\mathbf{Y}\!_N|\boldsymbol{\theta}) \pi(\boldsymbol{\theta})}\left[\|\boldsymbol{\theta}  - \hat{\varTheta}_{\boldsymbol{\alpha}}^{\mathcal{M}} (\mathbf{Y}\!_N)\|^2\right],
\end{equation}
\endgroup
as $P,M \to \infty$, where the expectation is with respect to the joint distribution $p(\mathbf{Y}\!_N|\boldsymbol{\theta}) \pi(\boldsymbol{\theta})$. The solution to the right hand side of (\ref{eq:BayesEstimator}) is conditional mean estimator of $\boldsymbol{\theta}$; namely
\begin{equation}
\mathbf{E} \left[\boldsymbol{\theta}\, |\cdot \right] = \int_\Theta \boldsymbol{\theta} \,   p(\boldsymbol{\theta} | \cdot) d \boldsymbol{\theta}.
\label{eq:exp_posterior}
\end{equation}
If this function can be realized using the chosen RNN, then for $P$ and $M$ chosen large enough, the DeepBayes Estimator $\hat{\varTheta}_{\hat{\boldsymbol{\alpha}}_{(P,M)}}^{\mathcal{M}}(\cdot; \beta)$ will perform close to the conditional mean estimator defined in \eqref{eq:exp_posterior}.

\subsection{An example of our estimation procedure: Linear Toy Model}\label{sec:LM_example}
Suppose that $\mathcal{M}$ is a set of finite impulse response (FIR) models of a order $L$, given as
\begin{equation}
y_{k} = \sum_{\ell=1}^{L} \theta_{l}u_{k - \ell + 1} + v_{k}, \hspace{0.1 in} 
\label{eq:LM_equation}
\end{equation}
where $ u_k = 0 \text{ for } k = -(L-1), \ldots, 0$, $u_k \sim \mathcal{U}\left[0,1\right]$ for $k \geq 1$ (independent over $k$), $v_k \sim \mathcal{N}(0,\lambda_v)$ is  unobserved measurement noise, and  $\boldsymbol{\theta} = [\theta_1 \ldots \theta_{L}]^\top$. Then for $N$ observations, we can write the input signal in a matrix form as:
\begin{equation}\label{eq:LM_Input_Matrix}
\boldsymbol{\Phi} := \begin{bmatrix}
u_1 & 0 &  \hdots & 0 \\
u_2 & u_1 &  \hdots & 0 \\
\vdots & \vdots & \ddots  & \vdots \\
u_{N} & u_{N-1} &  \hdots & u_{N-L+1} \\
\end{bmatrix}
\end{equation}
Using \eqref{eq:LM_Input_Matrix},  we can write \eqref{eq:LM_equation} as
\begin{equation}
    \mathbf{Y}\!_{N} = \boldsymbol{\Phi} \boldsymbol{\theta} + \mathbf{v}\!_N,
\label{eq:LM_model_vector_equation}
\end{equation}
where $\mathbf{Y}\!_{N} = \left[ y_1, \ldots, y\!_N \right]^{\top}, \mathbf{v}\!_N = \left[ v_1, \ldots, v\!_N \right]^{\top}$. 
We assume that the unknown parameter vector $\boldsymbol{\theta}$  is a random vector, drawn from the prior distribution $\pi\left(\boldsymbol{\theta}\right)$. The simulated output signals are then given by 
\begin{equation}\label{eq:linear_model_output signals}
    \mathbf{Y}\!_{N}(\boldsymbol{\theta}_p,m) = \boldsymbol{\Phi} \boldsymbol{\theta}_p + \mathbf{v}\!_N(m),
\end{equation}
where $\boldsymbol{\theta}_p \sim \pi\left(\boldsymbol{\theta}\right)$, with $p, m$ denoting indices for the realizations of $\boldsymbol{\theta}$ and the noise sequence, respectively. To gain some insight into our proposed procedure, let us constrain the estimator function $\hat{\varTheta}_{\hat{\boldsymbol{\alpha}}_{(P,M)}}^{\mathcal{M}}$ to the set of linear maps and the prior $\pi$ to be Gaussian. In this case the estimator function can be represented by a matrix $\mathbf{A} \in \mathbb{R}^{d\times N}$ to be learned by solving the optimization problem
\[
\min_{\mathbf{A}} \sum_{p=1}^P\sum_{m=1}^M \| \boldsymbol{\theta}_p - \mathbf{A}\, \mathbf{Y}\!_{N}(\boldsymbol{\theta}_p,m)\|^2.
\]
Now observe that this is a quadratic problem; using the vectorization operator we can rewrite this as
\[
\min_{\mathbf{A}} \sum_{p=1}^P\sum_{m=1}^M \|\boldsymbol{\theta}_p - (\mathbf{Y}\!_{N}^\top(\boldsymbol{\theta}_p,m) \otimes \mathbf{I}_N) \vec(\mathbf{A}) \|^2,
\]
where $\otimes$ denotes the Kronecker product and $\vec(\cdot)$ denotes the vectorization operation. 
Let $\mathbf{Q}_{(p,m)} := (\mathbf{Y}\!_{N}^\top(\boldsymbol{\theta}_p,m) \otimes \mathbf{I}\!_N)$ and $\boldsymbol{\alpha}:=\vec(\mathbf{A})$. The problem is then
\begin{equation*}\label{eq:lsproblem}
\min_{\boldsymbol{\alpha}} \sum_{p=1}^P\sum_{m=1}^M \left(\boldsymbol{\theta}_p - \mathbf{Q}_{(p,m)}\boldsymbol{\alpha}\right)^\top \left(\boldsymbol{\theta}_p - \mathbf{Q}_{(p,m)}\boldsymbol{\alpha}\right),
\end{equation*}
which has the closed-form solution
\begin{equation*}\label{eq:lssolution}
\hat{\boldsymbol{\alpha}}_{(P,M)} = \left(\sum_{p,m} \mathbf{Q}^\top_{(p,m)}\mathbf{Q}_{(p,m)} \right)^{-1} \sum_{p,m} \mathbf{Q}^\top_{(p,m)} \boldsymbol{\theta}_p.
\end{equation*}
It then follows that
\[
\hat{\mathbf{A}}_{(P,M)} = \left( (\vec(\mathbf{I}_d^\top)) \otimes \mathbf{I}\!_N\right) \left( \mathbf{I}\!_N \otimes \hat{\boldsymbol{\alpha}}_{(P,M)} \right),
\]
and for a test data set $\mathbf{Y}\!_{N, test}$ the estimator is given by
\begin{equation}\label{eq:optimized_A_hat}
\hat{\boldsymbol{\theta}}_{(P,M)} = \hat{\mathbf{A}}_{(P,M)}\mathbf{Y}\!_{N,test}.
\end{equation}
In fact, due to the independence of the sampled realizations,
a direct application of the law of large numbers yields that
\[
\hat{\mathbf{A}}_{(P,M)} \to \mathbf{A}_{\infty} : = \mathbf{R}_\theta\boldsymbol{\Phi}^\top (\boldsymbol{\Phi} \mathbf{R}_\theta \boldsymbol{\Phi}^\top + \lambda_v \mathbf{I}\!_N)^{-1},
\]
as $P, M \to \infty$, where 
\begin{equation}
\mathbf{R}_\theta := \mathbf{E}_{\pi\left(\boldsymbol{\theta}\right)}\left[\boldsymbol{\theta} \boldsymbol{\theta}^{\top}\right] = \lim_{P\to \infty}\frac{1}{P} \sum_{p=1}^P \boldsymbol{\theta}_p \boldsymbol{\theta}_p^\top,
\label{eq:R_zero_mean}
\end{equation}
and $\lambda_v$ is the variance of the noise process $v_{k}$.  
 Substituting $\hat{\mathbf{A}}_{(P,M)}$ in \eqref{eq:optimized_A_hat}, we get
\begin{equation*}\label{eq:asymp_theta_hat}
\boldsymbol{\hat{\theta}}_{(P,M)} \to \mathbf{R}_\theta\boldsymbol{\Phi}^\top (\boldsymbol{\Phi} \mathbf{R}_\theta \boldsymbol{\Phi}^\top + \lambda_v \mathbf{I}\!_N)^{-1}\mathbf{Y}\!_{N, test}.
\end{equation*}
In case the mean value of samples $\boldsymbol{\theta}_p$ is different from zero, the required estimator has the form $\mathbf{A}\left(\cdot\right) + \mathbf{b}$. Analogous to \eqref{eq:optimized_A_hat}, the estimator is then
\begin{equation}\label{eq:optimized_Ab_hat}
\hat{\boldsymbol{\theta}}_{(P,M)} = \hat{\mathbf{A}}_{(P,M)}\mathbf{Y}\!_{N,test} + \hat{\mathbf{b}}_{(P,M)}
\end{equation}
It can be shown that as $P, M \to \infty$ 
\begin{equation}
\begin{aligned}
\hat{\mathbf{A}}_{(P,M)} &\to \mathbf{A}_{\infty} := \mathbf{R}_\theta\boldsymbol{\Phi}^\top (\boldsymbol{\Phi} \mathbf{R}_\theta \boldsymbol{\Phi}^\top + \lambda_v \mathbf{I}\!_N)^{-1}, \\
\hat{\mathbf{b}}_{(P,M)} &\to \mathbf{b}_{\infty} :=
\left(\mathbf{I}_d - \mathbf{A}_{\infty}\boldsymbol{\Phi}\right)\boldsymbol{\mu}_{\theta}, \\
\end{aligned}
\label{eq:asymp_estimates_bias}
\end{equation}
where similar to \eqref{eq:R_zero_mean} $\boldsymbol{\mu}_{\theta} = \lim\limits_{P\to\infty} \frac{1}{P}\sum_{p=1}^P \boldsymbol{\theta}_p$, and
\begin{equation*}
    \mathbf{R}_\theta := \mathbf{E}_{\pi\left(\boldsymbol{\theta}\right)}\left[\left(\boldsymbol{\theta} - \boldsymbol{\mu}_{\theta}\right)\left(\boldsymbol{\theta} - \boldsymbol{\mu}_{\theta}\right)^{\top}\right].
\label{eq:R_mu}
\end{equation*}
The required asymptotic estimator using \eqref{eq:asymp_estimates_bias} is obtained as:
\begin{equation}
\begin{aligned}
\boldsymbol{\hat{\theta}}_{(P,M)} \to \boldsymbol{\theta}_{\infty} = \mathbf{A}_{\infty} \mathbf{Y}\!_{N, test} + \mathbf{b}_{\infty}. \\
\end{aligned}    
\label{eq:asymp_theta_hat_Ab}
\end{equation}
Since the prior on $\boldsymbol{\theta}$ is Gaussian, this means that the obtained estimator is asymptotically equivalent to the conditional mean \cite{pillonetto2014kernel}. This is illustrated in Section \ref{sec:linear_toy_model_results} using a simple simulation experiment.

We remark here that FIR models are quite simple and in practice we will need to work with more complicated models that are nonlinear in the input and have nonlinear parameterizations. Furthermore, the prior distributions used are usually non-Gaussian and often may have finite support, or infinite only on one side. In these cases, the posterior mean that we try to approximate will be nonlinear in the data. Therefore we need a more flexible class of functions $\hat{\varTheta}_{{\boldsymbol{\alpha}}_{(P,M)}}^{\mathcal{M}}$ and parameterization, instead of just a linear map as we considered in the FIR case; for which the purpose was to illustrate the method in a simple setting. 

\section{Simulation experiments}
\label{sec:simulations}
In this section, we describe the experimental setup and training strategy for the proposed DeepBayes method and discuss our obtained result. We call the DeepBayes methods utilising GRU and LSTM architectures as DeepBayes-GRU and DeepBayes-LSTM, respectively. We use the RNN architectures operating in a many-to-one mode, i.e. we train RNNs to process a sequence of inputs and produce a one output vector at the final time step as described in Section \ref{sec:deepRNN}. 

We conduct a first experiment to demonstrate our proposed approach on the linear toy model, as described in Section \ref{sec:LM_example}. Next, we evaluate our method on two different variations of a popular, nonlinear state space model, and finally also on a real-world, coupled electric drive model \cite{wigren2017coupled}. We compare the performance with state-of-the-art methods based on ML like PSEM \cite{schon2011system}, CPF-SAEM \cite{lindsten2013efficient}, and also with a conditional mean estimator using Metropolis-Hastings sampling (CME-MH) \cite{ninness2010bayesian}. 

The ML-based methods and the Markov chains of the CME-MH are all initialized in the neighbourhood of the true values of the parameters. This is to improve convergence and avoid bad local solutions. We describe the details regarding the tuning of the MH algorithm in Appendix \ref{sec:Appendix:A}. The code is available at: \texttt{github.com/anubhabghosh/ParamEstimation}.

We use the MSE of the parameter estimate for evaluating the performance of the DeepBayes estimator. Assuming the true value of the unknown parameter as $\boldsymbol{\theta}_{\circ}$, we generate the test data set $D\!_{N, test}^{(\kappa)}\left(\boldsymbol{\theta}_{\circ}\right)$ using the input signal $\lbrace u_k \rbrace$ and $\boldsymbol{\theta}_{\circ}$. $D\!_{N, test}^{(\kappa)}\left(\boldsymbol{\theta}_{\circ}\right) (\kappa = 1, 2, \ldots, K_{test})$ consists of $K_{test}$ output signals in total, and the MSE is computed using a Monte Carlo approximation according to
 \begingroup\makeatletter\def\f@size{9}\check@mathfonts
\begin{equation}\label{eqn:mse_risk_MSE}
    \!\!{\text{MSE}}(\boldsymbol{\theta}_\circ) \!:= \frac{1}{K_{test}}\!\sum_{\kappa=1}^{K_{test}} \|\boldsymbol{\theta}_\circ - \hat{\varTheta}_{\hat{\boldsymbol{\alpha}}_{(P,M)}}^{\mathcal{M}}(D\!_{N, test}^{(\kappa)}(\boldsymbol{\theta}_\circ))  \|^2
\end{equation}
\endgroup
where
$\hat{\varTheta}_{\hat{\boldsymbol{\alpha}}_{(P,M)}}^{\mathcal{M}}$ denotes the learned estimator using the DeepBayes method, with optimized parameters given by $\hat{\boldsymbol{\alpha}}_{(P,M)}$. We use the same test data set $D\!_{N, test}^{(\kappa)}\left(\boldsymbol{\theta}_{\circ}\right)$ for evaluating the competing methods - PSEM, CPF-SAEM and CME-MH using \eqref{eqn:mse_risk_MSE}. Note that while the other methods require parameter initialization close to the true values of the parameters, DeepBayes does not have such a requirement.

\subsection{Experiments on a linear toy model}\label{sec:linear_toy_model_results}
We consider the FIR model defined in \eqref{eq:LM_model_vector_equation}, with a filter order $L=2$, given by
\begin{equation*}
y_k = \theta_1 u_{k} + \theta_2 u_{k-1} + v_{k}, \hspace{0.1 in} u_0 = 0,
\end{equation*}
where the true parameters are assumed to be $\theta_1 = \theta_2 = 0.7$, each $u_k \sim \mathcal{U}\left[0, 1\right]$ (independent of $k$), and $v_{k} \sim \mathcal{N}(0,\lambda_v)$. The value of the variance $\lambda_v$ is selected as $0.09$, and assumed known when the estimator is trained. We assume that $N = 500$, and generate a test dataset $D\!_{N, test} = (\mathbf{Y}\!_{N}, \boldsymbol{U}\!_{N})$ using the true parameter values. We generate the training dataset $Z_{(P, M)}\left(\mathcal{M}\right) $ using a Gaussian prior distribution, more specifically  $\boldsymbol{\theta}_p \sim \mathcal{N}\left(\left[1, 1\right]^{\top} , \frac{1}{3}\mathbf{I}_2\right)$. The simulated output signals are then given by \eqref{eq:linear_model_output signals}
where $\boldsymbol{\Phi}$ is obtained using samples $u_k$ as in \eqref{eq:LM_Input_Matrix} and kept same during both training and testing.
\begin{table}[!htbp]
    \centering
    \caption{Variation of MSE computed using \eqref{eqn:mse_risk_MSE} between the optimized values of the parameter $\boldsymbol{\theta}$ using \eqref{eq:optimized_Ab_hat} and the asymptotic value of $\boldsymbol{\theta}$ using \eqref{eq:asymp_theta_hat_Ab} for the linear model. $P$ denotes number of realizations of the parameter vector, $M$ denotes the number of simulated output signals per realization of the parameter vector. All MSE values in the table are in the scale of $10^{-5}$.\\}
    \begin{tabular}{|c|c|c|c|c|}
        \hline
        \diagbox{P}{M} & 50 & 100 & 200 & 500 \\
        \hline
        \hline
        400 & 5.47 & 2.55 & 1.29 & 0.53 \\
        \hline
        500 & 4.26 & 1.90 & 1.04 & 0.44 \\
        \hline
        600 & 3.68 & 1.78 & 0.80 & 0.33 \\
        \hline
        1000 & 2.07 & 1.12 & 0.52 & 0.20 \\
        \hline
    \end{tabular}
    \label{tab:LM_results}
\end{table}
To study how the choice of user parameters affects the estimator, we use a range of $P$, the number of realizations of the parameter vector and $M$, the number of noise realizations for each sampled parameter vector. For each estimator, we compute estimates using \eqref{eq:optimized_Ab_hat} and \eqref{eq:asymp_theta_hat_Ab}
The results are shown in Table \ref{tab:LM_results}. We can see that for increasing values of $P, M$, the MSE between the estimates computed using \eqref{eq:asymp_theta_hat_Ab} and \eqref{eq:optimized_Ab_hat} indeed decreases, which corroborates our theory.

\subsection{Nonlinear state space models}
Similar to \cite{schon2011system, lindsten2013efficient}, we consider the following discrete-time nonlinear time series model:
 \begingroup\makeatletter\def\f@size{9}\check@mathfonts
\begin{equation}\label{eq:growthmodel_generic}
\!\!\!\mathcal{M} \!:\!
\begin{cases}
\begin{aligned}
x_{k+1} &= \theta_1 x_k + \theta_2\left( \frac{x_k}{\theta_{3}x_k^2+\theta_4}\right) + \theta_5 u_k + w_{k},\\
y_k &= \theta_6 x_k^2 + v_{k}, \\
\boldsymbol{\theta} &\sim \pi\left(\boldsymbol{\theta}\right),
\end{aligned}
\end{cases} 
\end{equation}
\endgroup
where $u_k = \cos(1.2 k)$, $w_{k} \sim \mathcal{N}(0, \theta_7)$ and $v_{k} \sim \mathcal{N}(0, \theta_8)$. We assume zero initial conditions and that $w_{k}$ and $v_{k}$ are independent over $k$ and of each other. The parameter vector to be identified is a subset of $\boldsymbol{\theta} = \left[\theta_1 \dots \theta_8 \right]^\top$. We consider two different variations of \eqref{eq:growthmodel_generic} and perform a comparative performance study of our DeepBayes estimation method. 
The first case is
 \begingroup\makeatletter\def\f@size{9.5}\check@mathfonts
\begin{equation}\label{eq:growthmodel_original}
\mathcal{M}_{1} \!:\!
\begin{cases}
\begin{aligned}
x_{k+1} &= \theta_1 x_k + \theta_2 \left(\frac{x_k}{x_k^2+1}\right) + 8 u_k + w_{k},\\
y_k &= \theta_6 x_k^2 + v_{k}, \\
\boldsymbol{\theta} &\sim \pi\left(\boldsymbol{\theta}\right),
\end{aligned}
\end{cases} 
\end{equation}
\endgroup
where $u_k = \cos(1.2 k)$, $w_{k} \sim \mathcal{N}(0, \theta_7)$ and $v_{k} \sim \mathcal{N}(0, \theta_8)$.
The parameter vector to be identified is $\boldsymbol{\theta} = \left[\theta_7, \theta_8 \right]^\top$ and the prior on the parameter is such that the parameters are mutually independent and uniformly distributed according to
\begin{equation}
\begin{aligned}
&\theta_7 \sim \mathcal{U}[0.1,1.5],
&\theta_8 \sim \mathcal{U}[\epsilon,1].
\end{aligned}
\label{eq:theta_limits}
\end{equation}
We assume the other parameters to be known and need not be identified: $\theta_1 = 0.5\,, \theta_2 = 25\,, \theta_3 = 1.0\,, \theta_4 = 1.0 \,, \theta_5 = 8.0 \,, \theta_6 = 1.0 $. 
The true parameter vector in this case is taken $
\boldsymbol{\theta}_\circ = \left[1, 0.1  \right]^\top
$.
The second case is
\begingroup\makeatletter\def\f@size{9}\check@mathfonts
\begin{equation}\label{eq:growthmodel_simplified}
\mathcal{M}_{2} \!:\!
\begin{cases}
\begin{aligned}
x_{k+1} &= \theta_{2} \left(\frac{x_k}{0.04x_{k}^2 + 1}\right) + u_k + w_{k},\\
y_k &= \theta_{6} x_k^2 + v_{k}, \\
\boldsymbol{\theta} &\sim \pi\left(\boldsymbol{\theta}\right),
\end{aligned}
\end{cases} 
\end{equation}
\endgroup 
where $u_k = \cos(1.2 k)$, $w_{k} \sim \mathcal{N}(0, \theta_{7})$ and $v_{k} \sim \mathcal{N}(0, \theta_{8})$.
The parameter vector to be identified is $\boldsymbol{\theta} = \left[\theta_2, \theta_6, \theta_7, \theta_8 \right]^\top$ where the prior $\pi\left(\boldsymbol{\theta}\right)$ is such that the parameters are mutually independent and uniformly distributed according to
\begin{equation}
\begin{aligned}
&\theta_{2} \sim \mathcal{U}[0, 1],     &\theta_{6} \sim \mathcal{U}[0.1, 2],\\        
&\theta_{7} \sim \mathcal{U}[\epsilon, 1],  &\theta_{8} \sim \mathcal{U}[\epsilon, 1].\\
\end{aligned}
\label{eq:theta_limits_simplified}
\end{equation}
In \eqref{eq:theta_limits_simplified}, $\epsilon>0$ is a very small strictly positive real number, since $\theta_7$ and $\theta_{8}$ are used for modeling variances. The values of the other parameters are assumed to be known and need not be identified: $\theta_1 = 0.0\,, \theta_3 = 0.04\,, \theta_4 = 1.0\,, \theta_5 = 1.0$. The true parameter vector is $\boldsymbol{\theta}_{\circ} = \left[0.7, 1.0, 0.1, 0.1\right]^{\top}$.

We choose to use $\mathcal{M}_1$ for estimating the variance parameters as it provides a familiar ground for comparison between existing ML-based methods in \cite{lindsten2013efficient, schon2011system}. For estimating a larger set of parameters, we use the simplified model set $\mathcal{M}_2$. We hope that the success of the DeepBayes method on \eqref{eq:growthmodel_simplified} will help motivate its application to further harder cases. 

\subsubsection{Training strategy and hyperparameter tuning}
\textit{Training details:} An epoch is defined as one complete pass through the training set, i.e. when the training algorithm has `seen' all the examples in the training set. The RNN architectures are trained using mini-batch gradient descent with a learning rate $\eta=0.001$, which is adaptively decreased in a \textit{step-wise} fashion by a factor of 0.9, with every one-third of the total number of epochs. This helps in preventing oscillations in training at the later stages. The optimizer used is Adam \cite{kingma2014adam}, and the training and inference algorithms have been written using PyTorch. We utilize the automatic differentiation library of PyTorch in computing the gradients for backpropagration \cite{paszke2019pytorch}.

\textit{Hyperparameter selection:}
Every RNN has a set of hyperparamters that need to be tuned in order to ensure desired performance. For our training purposes we choose to tune the following hyperparameters: number of hidden units ($n_{H}$) in each layer, number of hidden layers ($n_{l}$) and the number of hidden nodes in the final dense layer ($n_{z}$) prior to the output.
For tuning the hyperparameters, we monitor the validation loss at every epoch and enforce an early-stopping criterion to avoid overfitting \cite[Chap. 7]{goodfellow2016deep}. Early stopping is a very popular regularization method and introducing it into a training algorithm does not require any restrictions on the network structure or weights. Typically, it involves monitoring the change in the validation set performance for a specific number of iteration steps. The number of such iteration steps is often called \textit{patience}. If the performance does not improve beyond the patience parameter, the training is terminated and the model parameters and related metrics are recorded and saved. 
We describe our algorithm for early-stopping in Appendix \ref{sec:Appendix:B}.
\begin{table}[t]
    \centering
    \caption{Average MSE on the validation set for different configurations of the GRU architecture of a DeepBayes-GRU method. The validation MSE values are tabulated based on the \textit{best-saved} model for the task of estimating the variance parameters in $\mathcal{M}_1$. The chosen configuration for further experiments is highlighted.}
    \vspace{1em}
    \begin{tabular}{|c|c|c|c|c|c|c|}
         \hline
         S.No. & $n_{l}$ & $n_{H}$ & $n_{z}$ & Tr. MSE & Val. MSE \\
         \hline
         \hline
         1 & 1 & 30 & 32 & 0.00649 & 0.00675 \\
         \hline
         2 & 1 & 30 & 40 &  0.00569 &	0.00606 \\
         \hline
         \textbf{3} & \textbf{2} & \textbf{30} & \textbf{32} & \textbf{0.00505}	& \textbf{0.00558} \\
         \hline
         4 & 2 & 30 & 40 &  0.00554	& 0.00588\\
         \hline
         5 & 1 & 40 & 32 & 0.00534	& 0.00578 \\
         \hline
         6 & 1 & 40 & 40 &  0.00561 &	0.00600\\
         \hline
         7 & 2 & 40 & 32 & 0.00520	& 0.00576 \\
         \hline
         8 & 2 & 40 & 40 &  0.00517 &	0.00577 \\
         \hline
         9 & 1 & 50 & 32 & 0.00612	& 0.00627 \\
         \hline
         10 & 1 & 50 & 40 & 0.00595 &	0.00626 \\
         \hline
         11 & 2 & 50 & 32 & 0.00484 &	0.00584 \\
         \hline
         12 & 2 & 50 & 40 &  0.00443 &	0.00575 \\
         \hline
         13 & 1 & 60 & 32 & 0.00517 &	0.00595 \\
         \hline
         14 & 1 & 60 & 40 &  0.00511 &	0.00577 \\
         \hline
         15 & 2 & 60 & 32 & 0.00373 &	0.00631 \\
         \hline
         16 & 2 & 60 & 40 &  0.00334 &	0.00670 \\
         \hline
         \hline
    \end{tabular}
    \label{tab:val_hyp_selection_gru}
\end{table}
\subsubsection{Estimating the variance parameters: $\theta_7, \theta_8$ in $\mathcal{M}_1$}\label{sec:results_variances}
We generate the synthetic dataset $Z_{(P,M)}$ by uniformly sampling values of $[\theta_7, \theta_8]$, as defined in \eqref{eq:theta_limits}. We sample $P=500$ realizations of the parameter vector and $M=50$ output signals per realization. Each generated output signal is of length $N=200$. We split $Z_{(P,M)}$ into training and validation in the ratio $0.75 \colon 0.25$. The parameters of the RNN are learnt using the training set and hyperparameter tuning is done using the validation set. The results of hyperparameter tuning related to estimating the variance parameters of $\mathcal{M}_1$ task for a DeepBayes-GRU method are shown in Table \ref{tab:val_hyp_selection_gru}. We choose the configuration from Table \ref{tab:val_hyp_selection_gru} that provides the smallest validation error. Then, we proceed to retrain the model using a larger training data set consisting of $P=2000$ realizations with $M=500$ output signals generated per realization. For testing, we use the same value of $\boldsymbol{\theta}_{\circ}$ as $\left[1, 0.1\right]^\top$,
and generate $100$ output signals to form the test data set $D\!_{N, test}$. We compute parameter estimates for each output signal in $D\!_{N, test}$, then compare with the true value of the variance parameters taken from $\boldsymbol{\theta}_{\circ}$ using \eqref{eqn:mse_risk_MSE}. 

\begin{figure}[t]
    \centering
    \includegraphics[width=0.5\textwidth]{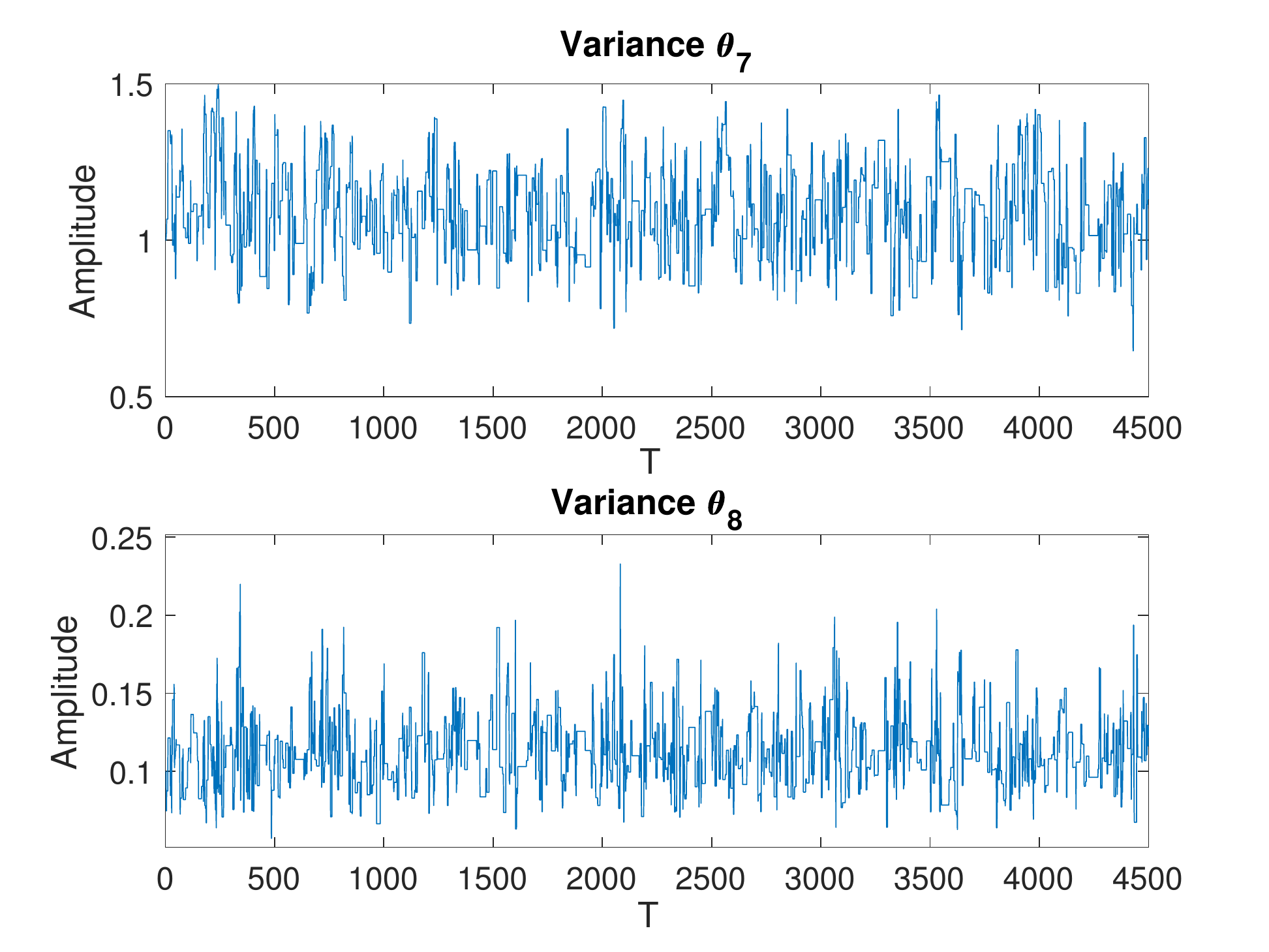}\vspace{-1em}
    \caption{The Markov chain of the Metropolis Hastings based estimates of the unknown parameters. The true value of the variances $\theta_7, \text{ and } \theta_8$ were 1.0 and 0.1 respectively. The plots indicate convergence of the Markov chains.}
    \label{fig:MH_variances}
\end{figure}

\begin{figure}[!htbp]
    \centering
    \includegraphics[width=0.5\textwidth]{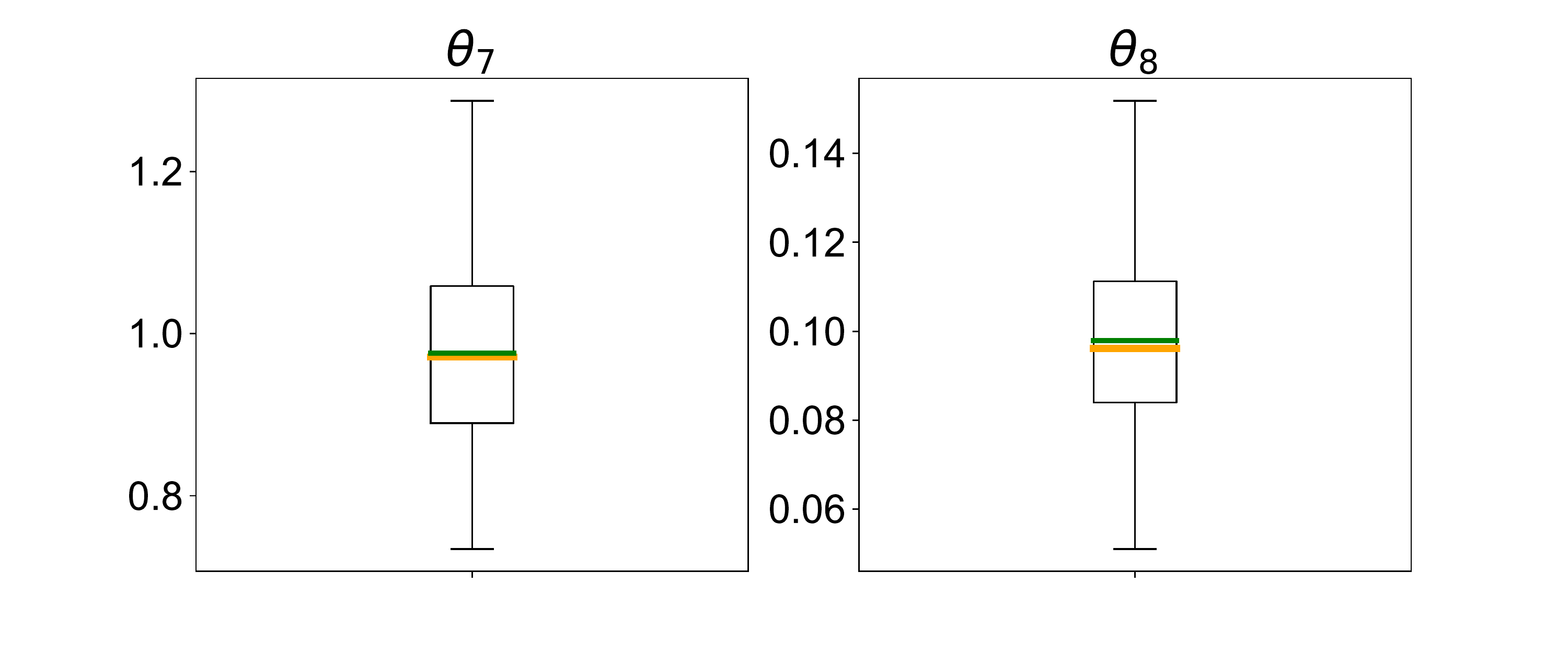}\vspace{-1em}
    \caption{Box-plot for estimates of the variances $\theta_7, \theta_8$ in $\mathcal{M}_1$ for a test set consisting of 100 samples using the DeepBayes-GRU method. The median and mean estimates are shown in orange and green respectively.}
    \label{fig:gru_boxplot}
\end{figure}

\begin{figure}[!htbp]
    \centering
    \includegraphics[width=0.5\textwidth]{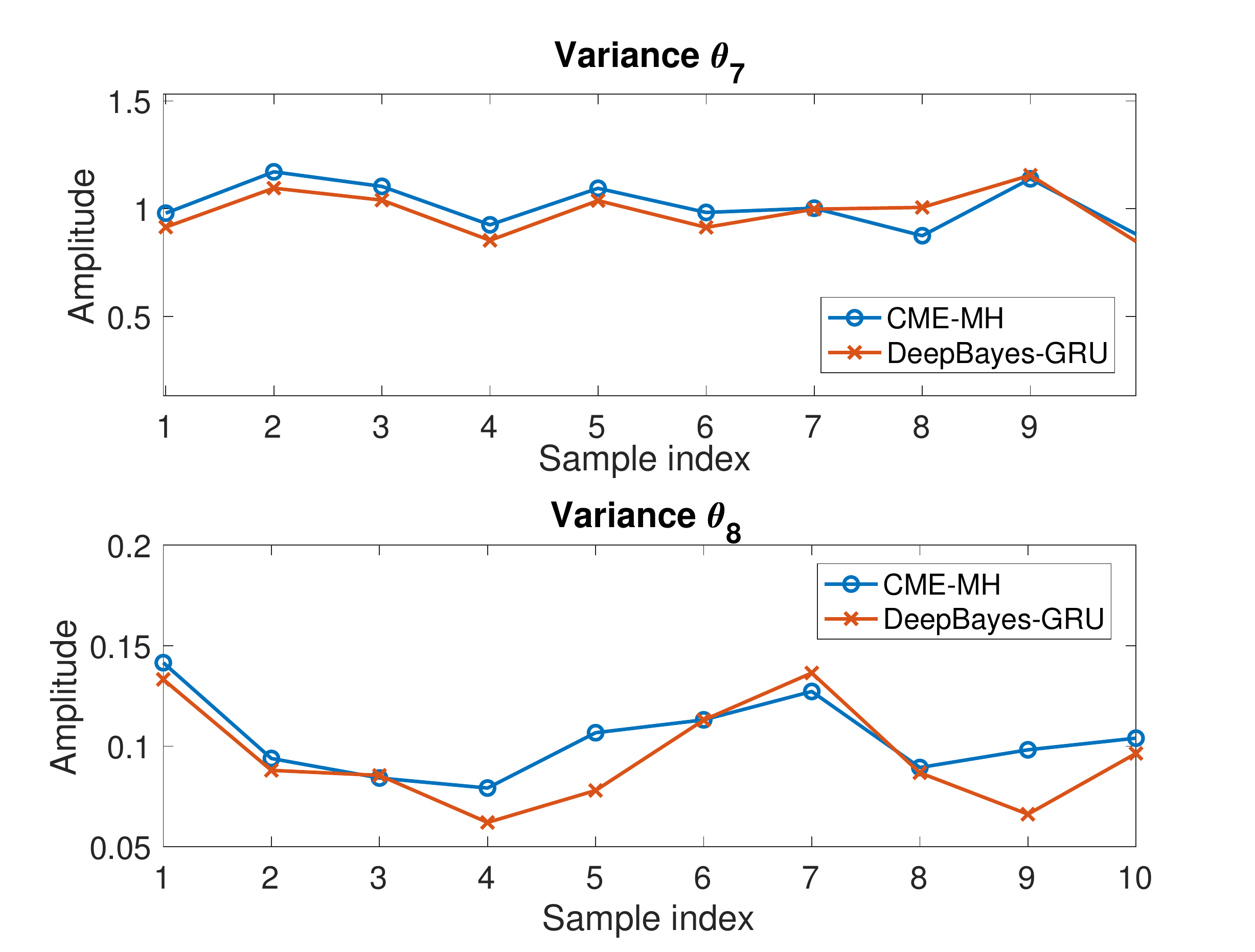} 
    \caption{A comparison of the estimates of variances $\theta_7, \theta_8$ of model $\mathcal{M}_1$ for 10 randomly chosen output signals from a test set consisting of 100 samples. The compared methods are CME-MH method (blue circles) and our proposed DeepBayes-GRU method (red crosses) . The x-axis represents indices of the randomly chosen output signals. The true values of $\theta_7, \theta_8$ were $1.0$ and $0.1$.}
    \label{fig:compare_preds}
\end{figure}

We repeat the same procedure for hyperaprameter selection using grid-search and testing for the DeepBayes-LSTM method.  The DeepBayes-LSTM method has more parameters compared to DeepBayes-GRU and this can lead to poorer generalization due to possible overfitting. We compare the results with the CPF-SAEM method \cite{lindsten2013efficient} and the CME-MH method, both evaluated on the same test set. The results are shown in Table \ref{tab:gru_final_results_modified_var} where the DeepBayes-GRU method outperforms CPF-SAEM by leveraging the prior knowledge and modeling capability, and comes quite close to the CME-MH method. An example of the plots of the Markov chain for parameters $\theta_7, \theta_8$ in case of the CME-MH method is also shown in Fig. \ref{fig:MH_variances}, which shows that the Markov chains indeed converged around the true values. 
A box plot showing the estimates for the DeepBayes-GRU method for the variance parameters $\theta_7, \theta_8$ is shown in Fig. \ref{fig:gru_boxplot}. Furthermore, we also provide in Fig. \ref{fig:compare_preds}, a comparison between the individual estimates of the benchmark CME-MH method and the DeepBayes-GRU on a finite number of randomly chosen output signals from the test set. We see from Fig. \ref{fig:compare_preds} that even the estimates  of the CME-MH and DeepBayes estimators on randomly chosen individual output signals are quite close. This indicates that in this example, the DeepBayes estimator successfully approximates the CME function. 
\begin{table*}[!htbp]
    \centering
    \caption{Average MSE on the test set using the DeepBayes-GRU and DeepBayes-LSTM methods for the task of estimating variance parameters of $\mathcal{M}_1$. The performance is compared with the ML-based method CPF-SAEM \cite{lindsten2013efficient} and the conditional mean estimator using Metropolis-Hastings (CME-MH).\\}
    \vspace{1em}
    \begin{tabular}{|c|c|c|c|c|}
        \hline
        Metric & CPF-SAEM & CME-MH & DeepBayes-LSTM & DeepBayes-GRU \\
        \hline
        \hline
        Test MSE & 0.00932 & 0.00774 & $\mathbf{0.00982}$ & $\mathbf{0.00875}$ \\
        \hline
    \end{tabular}
    \label{tab:gru_final_results_modified_var}
\end{table*}

\subsubsection{Estimating all the parameters in $\mathcal{M}_2$}
In this experiment, we seek to estimate the unknown parameters $\theta_2, \theta_6, \theta_7, \theta_8$ for $\mathcal{M}_2$ described in \eqref{eq:growthmodel_simplified}. The training dataset $Z_{(P,M)}$ is generated using a similar strategy as described in the previous setup for the variances in Section \ref{sec:results_variances}. The hyperparameters for the GRU architecture are chosen using the performance on a validation set, similar to the setup in Table \ref{tab:val_hyp_selection_gru}. This results in a GRU architecture having $n_l=2, n_H=40, \text{ and  }n_z = 32$. 
An important remark here is that the performance of our estimator method depends on the size of the training set. Through experiments, we find that the influence of the number of realizations $(P)$ is more than that of the number of output signals per each parameter realization $(M)$. In Table \ref{tab:simpler_model_results_modified_theta_diff_M}, we demonstrate this behaviour. We find that with the increase in P, keeping M constant, the performance of the model improves significantly, with the best performance given by the training set having $P=2000, M=500$. 

Hence, we select this training set configuration and compare with the ML-based PSEM method described in \cite{schon2011system}. 
The results are shown in Table \ref{tab:simpler_model_results_modified_theta}. Also, we numerically compute 
the conditional mean estimate in \eqref{eq:exp_posterior} using the CME-MH method. All results are shown for the same test set, which is generated by fixing the parameters to the known true values and simulating the model $\mathcal{M}_2$ described in \eqref{eq:growthmodel_simplified} for $M=100$ times. The performance metric used is  \eqref{eqn:mse_risk_MSE}. 

\begin{table}[!htbp]
    \centering
    \caption{Average MSE computed using \eqref{eqn:mse_risk_MSE} on the test set using the DeepBayes-GRU method for estimating all parameters in $\mathcal{M}_2$, for different values of $P$, with $M$ being kept fixed at 500}
    \vspace{1em}
    \begin{tabular}{|c|c|c|c|c|c|}
        \hline
        Metric & P=500 & P=1000  & P=1500 & P=2000 \\
        \hline
        \hline
        Test MSE & 0.00879 & 0.00326 & 0.00293 & 0.00265 \\
        \hline
    \end{tabular}
    \label{tab:simpler_model_results_modified_theta_diff_M}
\end{table}
\begin{table}[!htbp]
    \centering
    \caption{Average MSE computed using \eqref{eqn:mse_risk_MSE} on the test set using CME-MH, PSEM and DeepBayes-GRU methods. The task is estimating all parameters in $\mathcal{M}_2$ and the GRU-based method was trained using a training set having P=2000, M=500}
    \vspace{1em}
    \begin{tabular}{|c|c|c|c|}
        \hline
        Metric & PSEM & CME-MH & DeepBayes-GRU \\
        \hline
        \hline
        Test MSE & 0.00341 & 0.00261 & $\mathbf{0.00265}$ \\
        \hline
    \end{tabular}
    \label{tab:simpler_model_results_modified_theta}
\end{table}
\begin{table}[!htbp]
    \centering
    \caption{Comparing the training and inference times (in seconds) among CME-MH, PSEM and DeepBayes-GRU methods. The task is estimating all parameters in $\mathcal{M}_2$ and the DeepBayes-GRU method was trained using a training set having P=2000, M=500}
    \begin{tabular}{|c|c|c|}
        \hline
        Method & Training & Inference \\
        \hline
        \hline
        PSEM  & \multicolumn{2}{c|}{$0.67 \times 10^{4}$} \\
        \hline
        CME-MH  & \multicolumn{2}{c|}{$9.3 \times 10^{4}$} \\
        \hline
        DeepBayes-GRU  & $\mathbf{1.14 \times 10^{4}}$ & $\mathbf{8.9 \times 10^{-2}}$ \\
        \hline
    \end{tabular}
    \label{tab:tr_inference_times}
\end{table}
Although the ML-based methods are asymptotically optimal, we see from the experimental results in Table \ref{tab:simpler_model_results_modified_theta}, that the DeepBayes method is capable of producing results that are better than the ML-based methods and even close to the CME-MH method. More importantly, this is achieved by training only on synthetically generated data and having no special initialization close to the true values (as opposed to the PSEM method). This hints at the greater applicability of such models to real-world scenarios, where often true measurements are scarce and mostly unavailable during training. 
    
Another important advantage is that once these estimators are trained, inference tasks using these models are quite quick and efficient, without involving any re-training of architectures. We show in Table \ref{tab:tr_inference_times}, that the DeepBayes-GRU method ensures significant savings in inference time compared to the PSEM and CME-MH methods, while having fairly comparable training times. We remark that the use of deep RNNs in DeepBayes estimators requires the use of GPU for faster training. The PSEM and CME-MH methods have been implemented using MATLAB and trained on a CPU with an Apple M1-chip and 4 processing cores, while the DeepBayes methods have been implemented in PyTorch and run on a single NVIDIA Tesla P100 GPU card.

\subsection{Real-world data}\vspace{-0.5em}
In this section, we describe some experimental results on real-world data. Our system of interest is a pair of coupled electric drives. It consists of two electric motors that are coupled to a pulley through a conveyor belt, with the pulley held by a spring \cite{wigren2017coupled}. This system is a part of an open-source, nonlinear dynamical system model used for bench-marking purposes. A continuous-time description of 
the system is given by the Wiener model \vspace{-0.5em}
 \begingroup\makeatletter\def\f@size{9.5}\check@mathfonts
\begin{subequations}\label{eqn:basicwiener}
\begin{align}[left = \mathcal{M}_{CD}: \empheqlbrace\,]
	X\left(s\right) &= \frac{K \alpha \omega_{0}^{2}}{\left(s + \alpha\right) \left(s^{2} + 2\xi\omega_{0}s + \omega^{2}_{0}\right)}U\left(s\right) \label{eq:basicwiener_state}\\
    y_{k} &= \vert x\left(t_{k}\right) \vert + v_{k} \label{eq:basicwiener_output}\\
    \boldsymbol{\theta} &= \left[K, \alpha, \omega_{0}, \xi, \lambda_{v} \right],\label{eq:basicwiener_theta}
\end{align}
\end{subequations}
\endgroup
where $X\left(s\right) = \mathcal{L}\left(x\left(t\right)\right), U\left(s\right) = \mathcal{L}\left(u\left(t\right)\right)$. Here, $x\left(t\right)$ denotes the pulley velocity measured as voltage and the input signal $u\left(t\right)$ represents the sum of the voltages applied to both the motors. The measurements are recorded at sampled time instants $t = t_{k}$, so that $y_{k}$ denotes the measured, rectified noisy output signal and $v_{k} \sim \mathcal{N}\left(0, \lambda_{v}\right)$ denotes  additive Gaussian noise with variance $\lambda_v$. The parameter $\alpha$ denotes the inverse time constant of the electric drives, $\xi$ denotes the damping related to the spring and belt, $\omega_{0}$ denotes the angular resonance frequency, and $K$ denotes the static gain.  In \cite{wigren2005recursive}, a method for estimating the unknown parameters of this model through a recursive prediction error method (RPEM) has been developed. The authors in \cite{wigren2017coupled} consider three examples of input pseudo-random binary signals (PRBS) with different  amplitude values $u_{PRBS} = \left[0.5, 1.0, 1.5\right]$. For our experiments, we only consider the PRBS input signal with an amplitude of $0.5$. 
\subsubsection{Experimental setup in \cite{wigren2017coupled}}
In \cite{wigren2017coupled}, a measured signal of duration 10 seconds is used, with a sampling time of 20 ms. The sampling period for the input signal is 100 ms.   
The model structure can be written in state-space form as
\begin{subequations}
\begin{align}
\begin{bmatrix}
\frac{dx_{1}\left(t\right)}{dt} \\
\frac{dx_{2}\left(t\right)}{dt} \\
\frac{dx_{3}\left(t\right)}{dt} \\
\end{bmatrix}
&= \begin{bmatrix}
0 & 1 & 0 & 0 \\
0 & 0 & 1 & 0 \\
\theta_{4} & \theta_{3} & \theta_{2} & \theta_{1} \\ 
\end{bmatrix} \begin{bmatrix}
x_{1}\left(t\right) \\
x_{2}\left(t\right) \\
x_{3}\left(t\right) \\
u\left(t\right) \\
\end{bmatrix} \label{eq:ce_drive_state_model}\\
y\left(t\right) &= \vert x_{1}\left(t\right) \vert + e\left(t\right)\label{eq:ce_drive_output_model}.
\end{align}
\label{eqn:ce_drive_model}
\end{subequations}
We can write \eqref{eq:ce_drive_state_model} equivalently as
\begin{equation}
X_{1}\left(s\right) = \frac{\theta_{1}}{s^{3} - \theta_{2}s^{2} - \theta_{3}s - \theta_{4}} U\left(s\right),
\label{eqn:X1s}
\end{equation}
where, $X_{1}\left(s\right) = \mathcal{L}\left(x_{1}\left(t\right)\right), U\left(s\right) = \mathcal{L}\left(u\left(t\right)\right)$. The unknown parameter vector with a {continuous-time} parameterization is $\boldsymbol{\theta}^{\prime} = \left[\theta_1, \theta_2, \theta_3, \theta_4 \right]^{\top}$. Comparing \eqref{eq:basicwiener_state} and \eqref{eqn:X1s}, we can see that:
 \begingroup\makeatletter\def\f@size{9}\check@mathfonts
\begin{equation}\label{eq:tf_mapping}
\begin{aligned}
&K = -\dfrac{\theta_1}{\theta_4},\hspace{0.1 in} \omega_{0} = \sqrt{\frac{-\theta_4}{\alpha}}, \hspace{0.1 in} \xi = \frac{-\theta_2 - \alpha}{2\omega_0} \\
&\alpha = \text{\texttt{real roots of eqn. }} \left(s^{3} - \theta_{2}s^{2} - \theta_{3}s - \theta_{4} = 0\right). \\
\end{aligned}
\end{equation}
\endgroup
We also choose to model the unknown, additive output disturbance $v_{k}$ as white Gaussian noise with an unknown variance $\theta_5$. An important point to note is that RPEM involves estimating the unknown parameters of a continuous-time model that has been discretized using the Euler method \cite{wigren2005recursive}. As pointed out in \cite{wigren2017coupled}, this does not take into account the effect of sampling on the model. In our experimental setup, we use the obtained estimates of the parameters in continuous time reported in \cite{wigren2017coupled} as prior information for our DeepBayes method. We furthermore seek to overcome the limitations of the RPEM estimation, by performing discretization using zero-order hold, thus ensuring that our obtained parameters are indeed representative of the identified result.  
\begin{figure*}[t]
\centering
\includegraphics[width=0.8\textwidth]{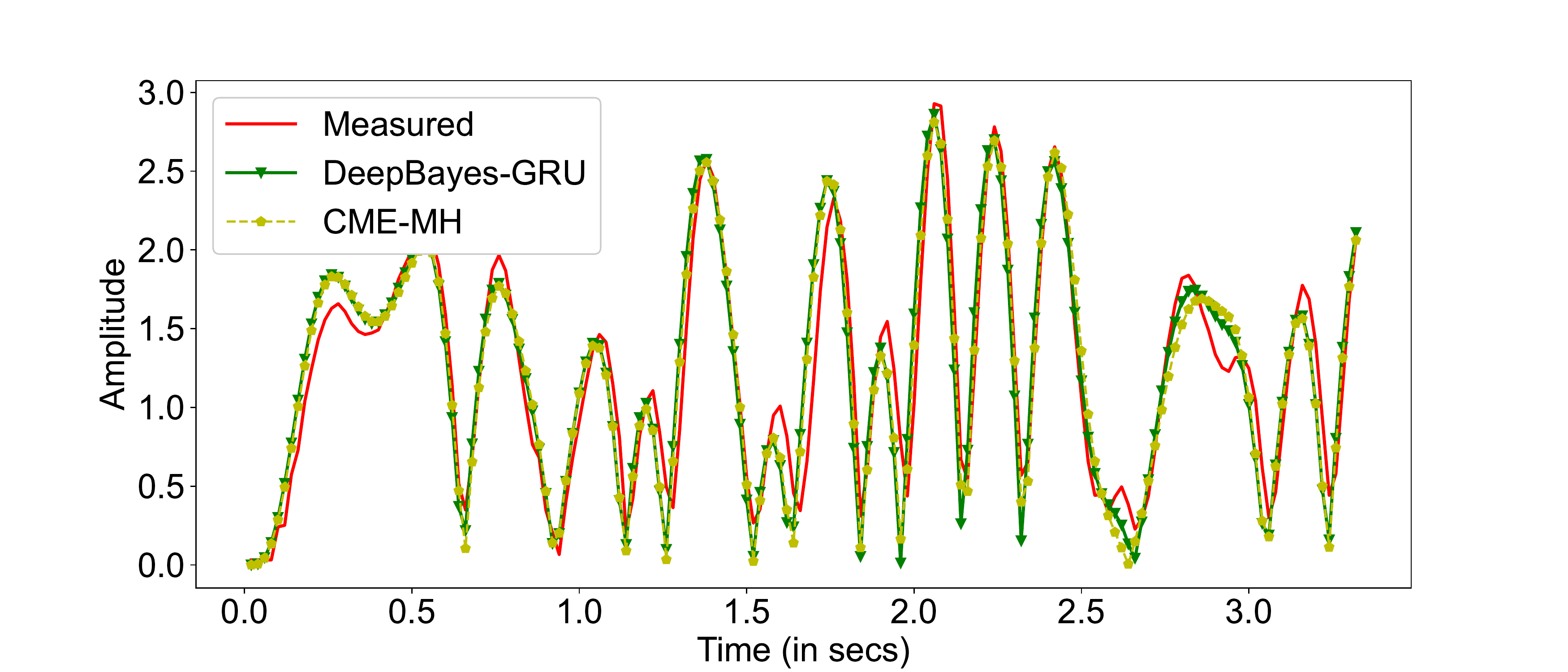}
\caption{Comparison of original output signal (in red) with the simulated output signals using the model structure and parameters found in \cite{wigren2017coupled} (in blue) and the simulated output signal using estimates from our trained DeepBayes-GRU method (in green) and CME-MH (in yellow). A shortened version of the original signal is shown here for ease of visualization.}
\label{fig:prbs_gru1}
\end{figure*}
\subsubsection{Dataset generation process}
For simulation purposes, we use the open source Python control systems library \footnote{Link:\texttt{https://python-control.readthedocs.io}}, which is similar to its MATLAB counterpart \cite{murray2018python}. The data set generation process for coupled electric drives model involves - choosing a reasonable prior to sample parameters and then simulating the model in \eqref{eqn:basicwiener} for generating output signals.

For the first step, we assume the prior distribution of our unknown parameters to be uniform, and seek to find a reasonable support for the prior. This is achieved using
a set of parameters $\boldsymbol{\theta}^{\prime}_{opt.} = \left[{\theta}_{opt., 1}, {\theta}_{opt., 2}, {\theta}_{opt., 3}, {\theta}_{opt., 4}\right]^{\top}$ obtained by solving the following nonlinear least squares problem 
\begin{equation*}
\label{eq:bruteforceoptim}
    \boldsymbol{\theta}^{\star} = \argmin\limits_{\boldsymbol{\theta}} \Vert \mathbf{Y}\!_{N, meas.} - \mathbf{Y}\!_{N}\left(\boldsymbol{\theta}\right) \Vert_{2},
\end{equation*}
where $\mathbf{Y}\!_{N, meas.}$ is the measured signal and $\mathbf{Y}\!_{N}\left(\boldsymbol{\theta}\right)$ is the simulated output signal of length $N$ using the model $\mathcal{M}_{CD}$ and the input $\mathbf{U}\!_{N}$ which is a PRBS input signal of length $N$ having amplitude $u_{PRBS} = 0.5$. 
In practice, the models are simulated using the command \texttt{lsim} in the control systems toolbox. We utilize the mapping between the parameters of the transfer functions in \eqref{eq:basicwiener_state}, \eqref{eqn:X1s} as shown in \eqref{eq:tf_mapping} to obtain $\boldsymbol{\theta}^{\prime}_{opt.}$ from $\boldsymbol{\theta}^{\star}$. Finally, to generate a synthetic dataset, we sample uniformly around 20 \% of $\boldsymbol{\theta}^{\prime}_{opt.}$ and for each sampled realization, we simulate the output signals as step responses of $\mathcal{M}_{CD}$ in \eqref{eqn:basicwiener} using $\mathbf{U}\!_{N}$. 
We assume $\lambda_v$ has a uniform prior $\left[\epsilon, 0.01\right]$. We train our DeepBayes-GRU method on a synthetic dataset consisting of $P=500$ realizations, with $M=1000$ output signals simulated per realization. Hyperparameters such as learning rate, early stopping patience, etc. are tuned according to performance on a subset of the synthetic dataset, chosen as the validation set. 
\begin{table}[!htbp]
    \centering
    \caption{Comparison of the sample sum of squared error between the measured output signal ($\mathbf{Y}\!_{N, meas.}$) and the simulated signal using parameter estimates from the DeepBayes-GRU method and the CME-MH method.\\}
    \begin{tabular}{|c|c|}
        \hline
        CME-MH & DeepBayes-GRU \\
        \hline
        \hline
        24.06 & $\mathbf{27.87}$ \\
        \hline
    \end{tabular}
    \label{tab:gru_final_results}
\end{table}
\vspace{-1em}
\subsubsection{Results}

We now describe the results of our method when applied to real-world data in Table \ref{tab:gru_final_results}. We compare our proposed DeepBayes-GRU method with the CME-MH method. Since we do not have `true' parameter values, the quality of each estimator is assessed by computing the sample, sum of squared error between the measured and the simulated output signal without disturbance using \eqref{eqn:ce_drive_model} and estimated parameters in each case. 
We obtain the reference estimate using the CME-MH method by employing a variation of Algorithm \ref{alg:mod_MH} where the log-likelihoods can be computed in an easier manner. Considering the model in \eqref{eqn:basicwiener}, we can see that given $x\left(t_k\right)$ and assuming $v_{k} \sim \mathcal{N}\left(0, \lambda_v\right)$, it follows that $y_{k} \sim \mathcal{N}\left(\vert x\left(t_k\right) \vert, \lambda_v\right)$. Hence, the log-likelihood of the signal $\mathbf{Y}\!_N$ (consisting of all samples $y_{k}$) is computed in \textit{closed form} using the fact that $\mathbf{Y}\!_N \sim \mathcal{N}\left(\vert \mathbf{x} \vert, \lambda_v \mathbf{I}\!_{N}\right)$. The resulting plot for the simulated output signals using the trained DeepBayes-GRU is shown in Fig. \ref{fig:prbs_gru1}. On visual inspection, we find that the simulated output signal using DeepBayes-GRU closely matches the measured output signal. Also from Table \ref{tab:gru_final_results}, we can see that the DeepBayes-GRU method performs quite satisfactorily compared to the CME-MH method.

\section{Conclusion and Future Work}\label{sec:conclusion}
In this contribution, we propose the DeepBayes method that leverages the modeling capability of deep RNNs for offline learning of estimators of nonlinear dynamical systems. We illustrate the asymptotic behaviour of our proposed estimators using a linear toy model. We then demonstrate that the estimation performance of DeepBayes is competitive with a state-of-the-art sampling based ML and Bayesian methods, with the distinct advantage of huge savings in inference time and no initialization issues. 

As part of future work, we would be interested in further investigating the effect of variable-length sequences. We would also like to investigate cases where the unknown parameters have different scales and analyse the relation between our method and recursive system identification. Another idea is to focus on learning prior distributions from data.

\section*{Appendix}
\appendix
\section{Numerically estimating the conditional mean estimate via Metropolis Hastings}\label{sec:Appendix:A}
We use a Markov Chain Monte Carlo method for numerically approximating the conditional mean estimator given by \eqref{eq:exp_posterior} in Section \ref{sec:theoretical_results}. In order to build the Markov chain, we use the MH algorithm \cite{hastings1970monte, chib1995understanding, robert1999metropolis}, and a particle filter implementation to approximate the log-likelihood \cite{schon2011system}. A reparameterization of the MH proposal distribution is done in order to sample effectively within the given prior. The MH algorithm is initialized using the true parameter $\boldsymbol{\theta}_{\circ}$ and the values of the proposal covariance $\Sigma_q$ are tuned to ensure convergence of the Markov chain. We also remark that we do not apply any thinning of the Markov chain and consider the average of the values of $\boldsymbol{\theta}_t$ in the Markov chain after $t \geq T_{\text{burn-in}}$, where $T_{\text{burn-in}}$ refers to the burn-in period \cite{robert1999metropolis}. The details of our implementation are shown in Appendix \ref{sec:Appendix:A}.

\begin{algorithm}[!htbp]
\caption{MH algorithm used in CME}
\label{alg:mod_MH}
\begin{algorithmic}
\State {\bfseries Input:} Dataset, $T_{\text{burn-in}}$, $T_{\text{reqd.}}, \boldsymbol{\theta}_{\circ}$
\State Initialize $\boldsymbol{\theta}^{(1)} \gets \boldsymbol{\theta}_{0}, \boldsymbol{\varphi}^{(1)} \gets g\left(\boldsymbol{\theta}^{(1)}\right)$, $T \gets T_{\text{burn-in}} + T_{\text{reqd.}} $. \\
In particular, $\varphi_{i} = g\left(\theta_{i}; a_{i}, b_{i}\right) = \frac{\log_{e}\left(\theta_{i} - a_{i}\right)}{\log_{e}\left(b_{i} - \theta_{i}\right)}$, where $\theta_{i} \sim \mathcal{U}\left[a_{i}, b_{i}\right]$, \\
$q\left(\mathbf{x} \vert \boldsymbol{\mu}, \Sigma_q\right) = \mathcal{N}\left(\mathbf{x} ; \boldsymbol{\mu}, \Sigma_q\right)$ (proposal distribution)
\For{$t=2$ {\bfseries to} $T$}
\State $\boldsymbol{\varphi}^{'} \sim q\left(\boldsymbol{\varphi}^{'} \vert \boldsymbol{\varphi}^{(t-1)}, \Sigma_{q} \right)$ \Comment{Sample a realization}
\State $\boldsymbol{\theta}^{'} \gets g^{-1}\left(\boldsymbol{\varphi}^{'}\right)$ \Comment{Obtain corresponding $\boldsymbol{\theta}$}
\State  $\boldsymbol{\Delta}_{\boldsymbol{\theta}^{'}} \gets \log_{e}\left(p\left(\mathbf{y} \vert \boldsymbol{\theta}^{'}\right)\right)$ 
\State $\boldsymbol{\Delta}_{\boldsymbol{\theta}^{(t-1)}} \gets \log_{e}\left(p\left(\mathbf{y} \vert \boldsymbol{\theta}^{(t-1)}\right)\right)$  
\begingroup\makeatletter\def\f@size{9.0}\check@mathfonts
\State $\begin{aligned} &\log A \gets \left(\boldsymbol{\Delta}_{\boldsymbol{\theta}^{'}} - \boldsymbol{\Delta}_{\boldsymbol{\theta}^{(t-1)}}\right) \\  
&+ \sum_{i} \left[\left(\boldsymbol{\varphi}^{'}_{i}  -  \boldsymbol{\varphi}^{(t-1)}_i\right) + 2\log\left(\frac{\exp\left(\boldsymbol{\varphi}^{(t-1)}_i\right) + 1}{\exp\left(\boldsymbol{\varphi}^{'}_i\right) + 1}\right)
\right] \end{aligned}$ \\\Comment{log of acceptance ratio}
\endgroup
\If{$\log A > \log u$}  \Comment{$u \sim \mathcal{U}[0, 1]$}
\State $\boldsymbol{\theta}^{(t)}, \boldsymbol{\varphi}^{(t)} \gets \boldsymbol{\theta}^{'}, \boldsymbol{\varphi}^{'}$ \\\Comment{Include new value in the chain}
\Else
\State $\boldsymbol{\theta}^{(t)}, \boldsymbol{\varphi}^{(t)} \gets \boldsymbol{\theta}^{(t-1)}, \boldsymbol{\varphi}^{(t-1)}$ \\\Comment{Copy previous value in the chain}
\EndIf
\EndFor
\State {\bfseries Output: } $\hat{\boldsymbol{\theta}}_{MH} \gets \frac{1}{T_{\text{reqd.}}} \sum_{t=T_{\text{burn-in}}+ 1}^{T} \boldsymbol{\theta}^{(t)}$
\end{algorithmic}
\label{algo:MH}
\end{algorithm}

\section{Description of our early stopping algorithm}
\label{sec:Appendix:B}
We describe the algorithm for early-stopping in Algorithm \ref{alg:validation_loss}. The algorithm monitors the relative change in the magnitude of the validation loss as a criterion to check convergence of the RNN training. In case the convergence criterion is satisfied, then the best model parameters, metrics are saved and the training is terminated. Else, the model parameters at the end of training is saved.  
\begin{algorithm}
\caption{Early stopping algorithm for regularization of the DeepBayes estimator}\label{alg:validation_loss}
\begin{algorithmic}
\State {\bfseries Input:} Patience ($p \geq 0$), tolerance $\Delta$, Initial model parameters $\boldsymbol{\alpha}^{(0)}$, Maximum number of epochs $N$
\State $\boldsymbol{\alpha} \gets \boldsymbol{\alpha}^{(0)}$
\State $i \gets 1$ \Comment{Epoch counter}
\State $j \gets 1$ \Comment{Counter for checking criterion}
\State $flag \gets False$ \Comment{Convergence flag}
\State $\mathcal{L}_v^{(0)} \gets \infty$ \Comment{$\mathcal{L}_v$ denotes the validation set error}
\State $\text{e}_{prev} \gets i-1$ 
\While{$i \leq N$}  \Comment{Training loop} \\
\State Update $\boldsymbol{\alpha}^{(i)}$ by running training algorithm using mini-batch gradient descent
\State  Compute $\mathcal{L}_v^{(i)}$ and set 
$\text{e}_{current} \gets i$  
\If{$\left[ \left(\frac{\abs{\mathcal{L}_v^{(i)} - \mathcal{L}_v^{(i-1)}}}{\abs{\mathcal{L}_v^{(i-1)}}} < \Delta \right)\right] ==$  \textbf{True} and $\text{e}_{prev} ==$ $\text{e}_{current} - 1$} 
    \State $j \gets j + 1$
    \If{$j == p$} \Comment{Criterion satisfied}
        \State $flag \gets True$
    \EndIf
    \State $\text{e}_{prev} \gets \text{e}_{current}$
\Else
    \State $j, flag \gets 1, False$
\EndIf
\State $i \gets i + 1$
\If{$flag == $\textbf{True}} \Comment{Save best parameters}
    \State $i^{\star}, \boldsymbol{\alpha}^{\star}, \mathcal{L}_v^{\star} \gets i, \boldsymbol{\alpha}^{(i)}, \mathcal{L}_v^{(i)}$
    \State Terminate while loop
\Else
    \State Continue training
\EndIf
\EndWhile
\end{algorithmic}
\end{algorithm}

\bibliographystyle{ieeetr}        
\bibliography{autosam}           
\end{document}